\pdfoutput=1

\documentclass[11pt]{article}

\usepackage[]{acl}

\usepackage{times}
\usepackage{latexsym}

\usepackage[T1]{fontenc}

\usepackage[utf8]{inputenc}

\usepackage{microtype}

\usepackage{graphicx} 
\usepackage{amsmath}
\usepackage{amssymb}
\usepackage{booktabs}
\usepackage{multirow}
\usepackage{enumitem}

\definecolor{cGreen}{RGB}{34,139,34}
\newcommand{\Revise}[1]{{\bf{\color{cRed}#1}}}
\definecolor{cRed}{RGB}{254,76,97}

\usepackage{soul}
\newcommand{\Del}[1]{}
\definecolor{cOrange}{RGB}{243,156,17}

\title{Modeling Complex Dialogue Mappings via Sentence Semantic Segmentation Guided Conditional Variational Auto-Encoder}

  \author{Bin Sun$^1$, Shaoxiong Feng$^1$, Yiwei Li$^1$, \\ \bf{Weichao Wang$^2$, Fei Mi$^2$, Yitong Li$^{2,3}$, Kan Li$^{1*}$}\\
  { $^1$School of Computer Science \& Technology, Beijing Institute of Technology }\\
  { $^2$ Huawei Noah's Ark Lab \quad  $^3$Huawei Technologies Ltd.} \\
  {\small\texttt{\{binsun,shaoxiongfeng,liyiwei,likan\}@bit.edu.cn} }\\
  {\small \texttt{\{wangweichao9,mifei2,liyitong3\}@huawei.com}}}

\begin{document}
\maketitle
\begin{abstract}
   Complex dialogue mappings (CDM), including one-to-many and many-to-one mappings, tend to make dialogue models generate incoherent or dull responses, and modeling these mappings remains a huge challenge for neural dialogue systems.
  To alleviate these problems, methods like introducing external information, reconstructing the optimization function, and manipulating data samples are proposed, while they primarily focus on avoiding training with CDM, inevitably weakening the model's ability of understanding CDM in human conversations and limiting further improvements in model performance.
  This paper proposes a Sentence Semantic \textbf{Seg}mentation guided \textbf{C}onditional \textbf{V}ariational \textbf{A}uto-\textbf{E}ncoder (SegCVAE) method which can model and take advantages of the CDM data.
  Specifically, to tackle the incoherent problem caused by one-to-many, SegCVAE uses response-related prominent semantics to constrained the latent variable.
  To mitigate the non-diverse problem brought by many-to-one, SegCVAE segments multiple prominent semantics to enrich the latent variables.
  Three novel components, Internal Separation, External Guidance, and Semantic Norms, are proposed to achieve SegCVAE.
  On dialogue generation tasks, both the automatic and human evaluation results show that SegCVAE achieves new state-of-the-art performance.
\end{abstract}

\section{Introduction}
\label{intro}

In open-domain conversations, complex dialogue mappings (CDM) between contexts and responses commonly exist in the real-world data, which bring considerable modeling challenges for neural dialogue models~\cite{FilteringData-Csaky-2019,Binsun-Sepacvae-ACL2021}: one-to-many mapping can cause models to generate incoherent responses, while many-to-one mapping makes the model produce non-diverse responses. 
For example, \texttt{CornellMovie}~\cite{cornellmovie} and \texttt{Opensubtitles}~\citep{opensubtitles2016} dialogue datasets contain 10.29\% (4.18\% + 6.11\%) and 9.10\% (4.79\% + 4.31\%) CDM data (one-to-many + many-to-one mappings) accordingly.
Many existing efforts tried identifying CDM and avoiding training on them to facilitate the dialogue learning.
\citeauthor{Attention-ThangLuong-2015,Persona-LiJiwei-2016} introduce external information to detach one-to-many pairs into one-to-one pairs, thus reducing the difficulty of model training.
Some works reconstruct the optimization functions, allowing model to learn from self-generated qualified responses instead of the ground-truth, thereby avoiding the directly training on many-to-one pairs~\citep{RLdialoguesys-LiJiwei-2016,ConverseGAN(AIM)-ZhangYizhe-2018,RL-P2BOT-Liu2020}.
Others train the model through filtered corpora, which usually contains few one-to-many and many-to-one dialogue pairs~\citep{FilterCoherence-Xu-2018,FilteringData-Csaky-2019,FilterConsistency-Akama-2020}.
For an instance, \citet{FilteringData-Csaky-2019} reported the improvement of a dialogue model with high entropy dialogue pairs (i.e. CDM) filtered out for training, which is consistent with our preliminary experiments in Table~\ref{tab:influence_of_complex_dialogue_mappings}.

\begin{table}[!t]
\small
\centering
\renewcommand\tabcolsep{3.5pt}
\begin{tabular}{lcccc}
  \toprule
  \textbf{Setting}          & \textbf{Distinct-3} & \textbf{BLEU} & \textbf{Emb.Aver.} & \textbf{Coherence}\\
  \midrule
  w. CDM   & \textbf{0.033} & 0.157 & 0.853 & 0.828 \\
  w/o. CDM & 0.028 & \textbf{0.192} & \textbf{0.859} & 0.828
  \\
  \midrule
  w. CDM   & \textbf{0.031} & 0.131 & 0.465 & 0.281 \\
  w/o. CDM & 0.027 & \textbf{0.149} & \textbf{0.469} & \textbf{0.282}\\
  \bottomrule
\end{tabular}
\caption{Preliminary experiments of Seq2Seq models trained with and without CDM on \texttt{CornellMovie} (up) and \texttt{Opensubtitles} (down).
}
\label{tab:influence_of_complex_dialogue_mappings}
\end{table}

\begin{figure}[!t]
\centering
\includegraphics[width=\linewidth]{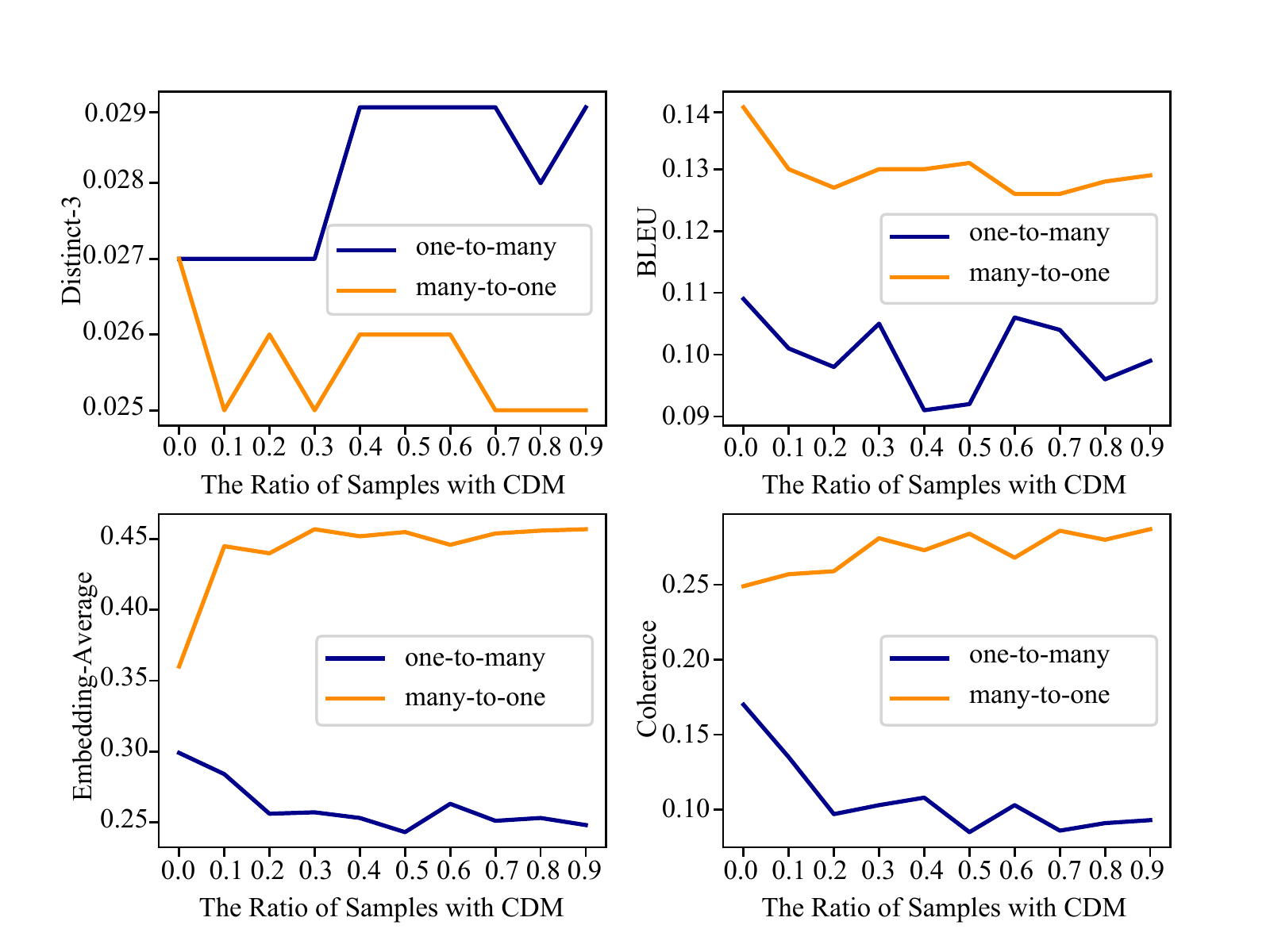}
\caption{Four metrics of Seq2Seq models fine-tuned by increasing one-to-many and many-to-one dialogue pairs.}
\label{fig-exp-invesrigation}
\end{figure}

Table~\ref{tab:influence_of_complex_dialogue_mappings} shows the comparison results of the same Seq2Seq dialogue model trained with/without CDM.
We can observe that the Seq2Seq trained without CDM improves the BLEU~\citep{bleu}, Emb.Aver.~\citep{embedding-16} and Coherence~\citep{coherence} but reduce the Distinct~\citep{distinct-16} (metrics detailed in Appendix~\ref{sect:dia-met}). Moreover, the gains on BLEU are big, but the gains on Emb.Aver. and Coherence are small.
This result proves the idea that reducing the CDM of the dataset is beneficial for increasing the scores of some automatic evaluation metrics. 

However, these methods simply ignore the CDM data (10\% of the dataset), and in this paper, we argue that these CDM dialogue pairs are still valuable for dialogue training.
To explore this, we conduct further experimental investigation by training two Sequence-to-Sequence dialogue models (Seq2Seq) \citep{Seq2Seq-ShangLifeng-2015} over the ``clean'' \texttt{Opensubtitles} dataset which does not contain any one-to-many or many-to-one pairs, respectively, and then we gradually introduce one-to-many/many-to-one pairs to fine-tune these models.
From Figure~\ref{fig-exp-invesrigation}, we observe that one-to-many and many-to-one dialogue pairs have conflicting effects on Distinct, Emb.Aver. and Coherence, which explains why simply removing them together yields smaller gains.
Therefore, instead of staying away from CDM, our primary study of interest is to enable model to effectively learn useful knowledge from these dialogue pairs while avoiding being affected by the disadvantages.

To achieve this goal, we take inspirations from Conditional Variational AutoEncoder (CVAE) based dialogue generation methods~\cite{CVAE(SPhred)-ShenXiaoyu-2017,kgCVAE-ZhaoTiancheng-2017,HVaeMN-ChenHongshen-2018,discrete-cvae-2019-emnlp,Binsun-Sepacvae-ACL2021}
and model the many-to-one and one-to-many from the latent space.
However, previous study shows that due to lack of the prior knowledge, latent variable hardly involves semantic relationships, resulting in semantically irrelevant responses~\cite{Binsun-Sepacvae-ACL2021}.
Therefore, we propose a Sentence Semantic Segmentation guided CVAE (SegCVAE),
using the sentence semantic segmentation to constrain the latent variable, which models the CDM naturally.

\begin{figure}[t]
\centering
\includegraphics[width=0.98\linewidth]{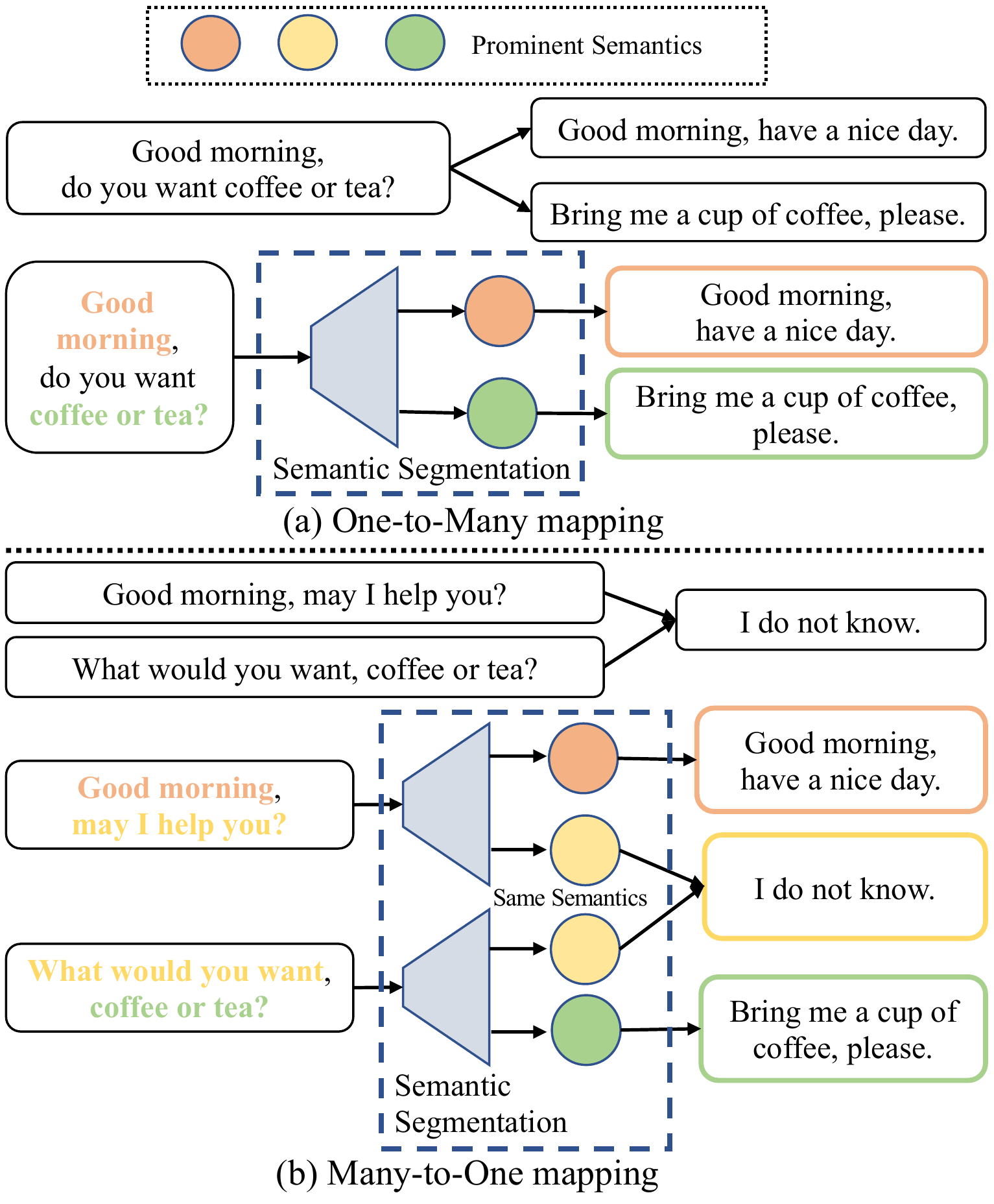}
\caption{The schematic of CDM and our primary idea for modeling CDM. (a) Multiple responses in an one-to-many mapping can disrupt model's ability to address the dialogue context. We associate different responses with the segmented different prominent semantics, so as to avoid the interference of multiple responses and to enhance the coherence. (b) The response in a many-to-one mapping has a high proportion in dataset, which deceives models into increasing the generation probability of it, reducing the diversity of generated responses. We promote the same prominent semantics to be associated with the same response, thus extending the response space to enhance the diversity.}
\label{fig-schematic-CDM-vertical}
\end{figure}

The complex and ambiguous context semantics can be reduced when segmented into multiple different sub-semantics, so that each sub-semantics may focus on different perspectives of the context.
We refer these sub-semantics to as prominent semantics, which can explain CDM naturally (see Figure~\ref{fig-schematic-CDM-vertical}):
When the semantics of a context being segmented into multiple prominent semantics, each of them corresponds to a response (i.e. one-to-many mapping); vice versa, when the prominent semantics is segmented by different contexts semantics, the same prominent semantics can correspond to the same response (i.e. many-to-one mapping).
To achieve this goal, we propose \textsc{Internal Separation} (IS) and \textsc{External Guidance} (EG) to model the prominent semantics together.
The IS extracts multiple different words from the context to obtain the prominent semantics.
The EG extracts the instructive words from the vocabulary to constrain the prominent semantics not far from the original semantics.
Furthermore, to make the prominent semantics capture the relationship with responses and latent variables, we propose \textsc{semantic alienation norm}, \textsc{semantic centralization norm}, and \textsc{semantic distillation norm} to regularise the learning of CVAE.

Our contributions are as follow:
\begin{itemize}[align = left, wide = 1pt, itemsep=2pt, parsep=2pt,topsep = 2pt ]
\item
We propose SegCVAE to model CDM through using sentence semantics segmentation (IS and EG) guided latent variables. SegCVAE constructs the relationships between multiple responses and multiple prominent semantics, thereby naturally explaining CDM.
Hence, prominent semantics can constrain latent variables to involve semantic relations when modeling CDM.
\item
We present \textsc{semantic alienation norm}, \textsc{semantic centralization norm}, and \textsc{semantic distillation norm} to regularize prominent semantics and facilitate semantic segmentation without supervised labels.
\item
We conduct extensive experiments to show the superior performance of SegCVAE in modeling CDM and dealing with the open-domain dialogue generation task.
\end{itemize}

\section{Related Work}
The open-domain dialogue generation has received dramatic attention recently \citep{Seq2Seq-Sutskever-2014,Seq2Seq-ShangLifeng-2015,Seq2Seq-Sordoni-2015}.
\citet{Seq2Seq-Sutskever-2014} identified that ``noisy'' data, including one-to-many and many-to-one dialogue pairs, can affect the performance of dialogue systems.
To address such ``noisy'' data, many methods have been proposed in recent years.
For instance, a large body of work on introducing external information for reducing the number of noisy data \citep{Attention-ThangLuong-2015,Persona-LiJiwei-2016,HRED-Dialogue-Serban-2016,kgCVAE-ZhaoTiancheng-2017,EmotionalTextImage-Huber-2018,FactKnowledge-Ghazvininejad-2018,CMHAM-TaoChongyang-2018,HVaeMN-ChenHongshen-2018,Regularizing-Feng-2020}, and a rich line of work reconstructs the objective function to avoid training models directly on such noisy data \citep{RLdialoguesys-LiJiwei-2016,GAN-GANAEL-Xu2017,RL-Seq2seqCo-Zhang2018,DPGAN-XuJingjing-2018,ConverseGAN(AIM)-ZhangYizhe-2018,PosteriorGan-FengShaoxiong2020,RL-P2BOT-Liu2020,NegativeTrain-He-2020,PanGuBot-feimi-2022,THINK-binsun-2022,NegDistill-yiweili-naacl-2022}. Others design a scoring approach to filter noisy data \citep{FilterCoherence-Xu-2018,FilteringData-Csaky-2019,FilterConsistency-Akama-2020,StopFiltering-yiweili-2022}; 

However, CDM data in human conversations impels valuable information that can help models generate better responses, and these methods cannot learn the valuable information of one-to-many and many-to-one dialogue pairs, nor can they make full use of the advantages of these data. 
For example, \citet{Persona-LiJiwei-2016} uses personal information to reduce the one-to-many dialogue pairs.\Del{They believed that different personal information with the same context will lead to different responses.}
The Reinforcement Learning based dialogue generation methods \citep{RLdialoguesys-LiJiwei-2016,RL-Seq2seqCo-Zhang2018} only require the generated response to get high reward rather than similar with the ground-truth, which means that some many-to-one dialogue pairs are ignored during training.
\citet{FilteringData-Csaky-2019} uses conditional entropy to assess the dialogue pairs, which easily filters one-to-many and many-to-one dialogue pairs.

In addition to the methods above, CVAE-based dialogue generation methods \cite{CVAE(SPhred)-ShenXiaoyu-2017,kgCVAE-ZhaoTiancheng-2017,HVaeMN-ChenHongshen-2018,discrete-cvae-2019-emnlp,DBLP:conf/emnlp/WangFWZ19,Binsun-Sepacvae-ACL2021} provide an idea to learn the essential knowledge of the one-to-many and many-to-one mappings.
They try to encode knowledge into a latent space, a posterior probability distribution, and a prior probability distribution.
By sampling latent variables, the model can easily generate multiple responses for one context.
We follow this rich line of work to explore their applicability in modeling CDM, and we propose new state-of-the-art SegCVAE in dialogue generation task.
Compared with the vanilla CVAE, SegCVAE uses sentence semantic segmentation to regularize and guide the latent variables, which avoids the gap between context and latent variables. 
Different from knowledge-guide CVAE, SegCVAE does not require additional information.
Meanwhile, SegCVAE uses the segmented prominent semantics instead of manually-created orthogonal vectors, which is more reasonable than SepaCVAE.

\section{SegCVAE}
\label{method}
SegCVAE is proposed to model CDM (including one-to-many and many-to-one mappings) through sentence semantic segmentation guided latent variables. 
As discussed above, different prominent semantics can be segmented from one context semantics, and similar prominent semantics can be segmented from different context semantics, which help latent variables learn the semantic relations, thus modeling one-to-many and many-to-one naturally.
In this section, we provide detailed descriptions of the proposed SegCVAE method.

\subsection{Overview}
SegCVAE uses multiple prominent semantics $(x_1, x_2, x_3, \ldots)$ to learn the probability distribution over response with latent variables, and $x_i$ denotes the representation of one prominent semantics. To train SegCVAE, we derive the \textit{Stochastic Gradient Variational Bayes} framework \citep{StochasticGradientVariationalBayes-kingma-2014,ELBO1-Sohn-2015,ELBO2-Yan-2016} and \textit{gradient blocking} trick \citep{Binsun-Sepacvae-ACL2021}:
\begin{align}
\label{eq:L(r,Ei)}
    \mathcal{L}(r,x^+) &= \max_{i=1,2,3,\ldots} \mathcal{L}(r,x_i),\\
    \nonumber \mathcal{L}(r,x_i) &=  \mathbb{E}_{q_{\phi}(z|r_e,x_i)}(\log p_{\Omega}(r|z, x_i)) \\
    &- KL(q_{\phi}(z|r_e,x_i)||p_{\theta}(z|x_i)),
\end{align}
where $q_{\phi}(z|r_e,x_i)$ and $p_{\theta}(z|x_i)$ are the recognition network and the prior network that used for sampling latent variable $z$, respectively. The $r_e=enc(r)$ is the semantic vector computed by model's encoder $enc$ based on the response $r$.
The $p_{\Omega}$ denotes the model's decoder, which generates the output token based on the conditional probability $p_{\Omega}(r|z, x_i)$.
Following the \textit{gradient blocking} trick, $x^+ \in (x_1, x_2, x_3, \ldots)$ denotes the prominent semantics vector that makes the variational lower bound largest, and only $\mathcal{L}(r,x^+)$ is used to optimize the model.

To obtain the prominent semantics $(x_1, x_2, \ldots)$, SegCVAE employs the \textsc{internal separation} (IS) and \textsc{external guidance} (EG).
To further capture the relationship among context, prominent semantics, and response, we propose three novel semantic norms: \textsc{semantic alienation norm}, \textsc{semantic centralization norm}, and \textsc{semantic distillation norm}.

\subsection{Internal Separation}
The IS processes sentences through multiple triggers and extracts multiple sets of different words, which can be used to compute different prominent semantics.
Each trigger consists of a convolution network $Conv$ and a dense network $Dense$.
The input of a trigger is an embedded matrix representation $\mathbf{C}$ of a context with a shape ($max\_clen, N$), where $max\_clen$ represents the maximum length of a context that can be received and $N$ is the dimension of the word-embedding.
The $\mathbf{C}$ is processed by $Conv$ whose kernel $K$ and stride $S$ are $(m, N, 1, chan)$ and $(1,1,1,1)$, respectively. The $chan$ is the number of channels of the convolution operation, and ($m$, $N$) denotes the shape of convolution kernel.
\begin{equation}
    \mathcal{F}_c = Conv(\mathbf{C}, K, S)
\end{equation}

After that, we get the semantic features $\mathcal{F}_c$.
We squeeze and transpose the $\mathcal{F}_c$ from $(max\_clen-m+1, 1, chan)$ to $(chan, max\_clen-m+1)$, and put it into the $Dense$. 
The weight of $Dense$ is $\mathcal{W}$ with a shape ($max\_clen-m+1, max\_clen$).

We use $SoftMax$ function to handle the last dimension of the input ($\mathcal{F}_c \cdot \mathcal{W}$). \begin{align}
\label{eq:Fd}
    \mathcal{F}_d = SoftMax(\mathcal{F}_c \cdot \mathcal{W})
\end{align}

Hence, the shape of $\mathcal{F}_d$ is $(chan, max\_clen)$, which represents the probability of words in the context of attention in different channels.
Then, we select the word with highest probability in each channel, which is processed by encoder $enc$ to extract certain semantic information.
However, this discrete process will hamper the optimization of model.
To ensure the gradient back-propagation, we introduce Gumbel SoftMax ($\mathbf{GS}$; \citet{gumbel-softmax-17}) to replace the $SoftMax$ (Eq.~\ref{eq:Fd}) and selection process:
\begin{align}
\label{eq:fdgs}
    & \mathcal{F}'_d = \mathbf{GS}(\mathcal{F}_c \cdot \mathcal{W}),
    \ \mathbf{GS}(\mathbf{Input})=\\
    \nonumber & \begin{pmatrix}
    \frac{e^{input_{11}/\tau}}{\sum^n_{k=1}e^{input_{1k}/\tau}}& \cdots &\frac{e^{input_{1n}/\tau}}{\sum^n_{k=1}e^{input_{1k}/\tau}}\\
    \vdots&\ddots&\vdots\\
    \frac{e^{input_{m1}/\tau}}{\sum^n_{k=1}e^{input_{mk}/\tau}}&\cdots &\frac{e^{input_{m1}/\tau}}{\sum^n_{k=1}e^{input_{mk}/\tau}}
    \end{pmatrix},
\end{align}
where $input_{ij} \in \mathbf{Input}$ and $\tau$ is the temperature parameter.
We control $\tau$ to be as small as possible, so that the output of $\mathbf{GS}$ is as close as possible to the result of $argmax(\mathcal{F}_d)$.
Thence, we can get the embedded matrix representation of extracted words $\mathbf{C}_{IS}=\mathcal{F}'_d \cdot \mathbf{C}$ with the shape of ($chan, N$).

Finally, we randomly initialize $\mathcal{M}$ trigger networks in IS to extract $\mathcal{M}$ embedded matrix representations ($\mathbf{C}^{1}_{IS}$, $\mathbf{C}^{2}_{IS}$, \ldots, $\mathbf{C}^{\mathcal{M}}_{IS}$) of different word-combinations from a context.

\subsection{External Guidance}
The EG is responsible for extracting instructive information from the outside of the sentence (i.e. the vocabulary) according to the context semantics.
To achieve this goal, we change the hyper-parameter of the dense network in the trigger defined in the previous section. The new weight matrix of the dense in EG is $\mathcal{W}'$, whose shape is changed from $(max\_clen-m+1, max\_clen)$ to $(max\_clen-m+1, vocab\_size)$, where $vocab\_size$ is the size of the vocabulary.
Hence, the results of the dense network denote the probability of words in the vocabulary of attention in different channels. Therefore, the output of EG is a matrix representation $\mathbf{V}_{EG}$ of $chan$ words in vocabulary related to the semantics of the input \Del{sentence}:
\begin{align}
    V_{EG} = \mathbf{GS}(\mathcal{F}_c \cdot \mathcal{W}') \cdot \mathcal{W}_{emb}
\end{align}
where $\mathcal{W}_{emb}$ is the word-embedding matrix whose shape is ($vocab\_size, N$).
Finally, we can also randomly initializes $\mathcal{M}$ new triggers in EG to extract $\mathbf{V}^{1}_{EG}$, $\mathbf{V}^{2}_{EG}$, $\ldots$, $\mathbf{V}^{\mathcal{M}}_{EG}$.
Therefore, the $\mathbf{C}_{IS}$ and the $\mathbf{V}_{EG}$ are used together to calculate multiple different prominent semantics of a context:
\begin{align}
    x_i = enc([\mathbf{C}_{IS}^{i}, \mathbf{V}^{i}_{EG}]) \, | \, i=1,2,\ldots,\mathcal{M},
\end{align}
where $enc$ denotes the model's encoder, $x_i$ represents $i$-th prominent semantics.

\subsection{Semantic Norms}

We consider self-supervise learning methods and propose \textsc{semantic alienation norm} ($\mathcal{L}_{san}$), \textsc{semantic centralization norm} ($\mathcal{L}_{scn}$), and \textsc{semantic distillation norm} ($\mathcal{L}_{sdn}$), to constrain the relations among the context, prominent semantics and response.
$\mathcal{L}_{san}$ and $\mathcal{L}_{scn}$ are responsible for promoting the multiple prominent semantics to be closely connected with the context on the basis of maintaining their own independence, which leverages the diversity and coherence of generated responses. $\mathcal{L}_{sdn}$ is used to facilitate the construction of semantic relations among prominent semantics.

\subsubsection{Semantic Alienation Norm}
\begin{figure}[!t]
  \centering
  \includegraphics[width=0.8\linewidth]{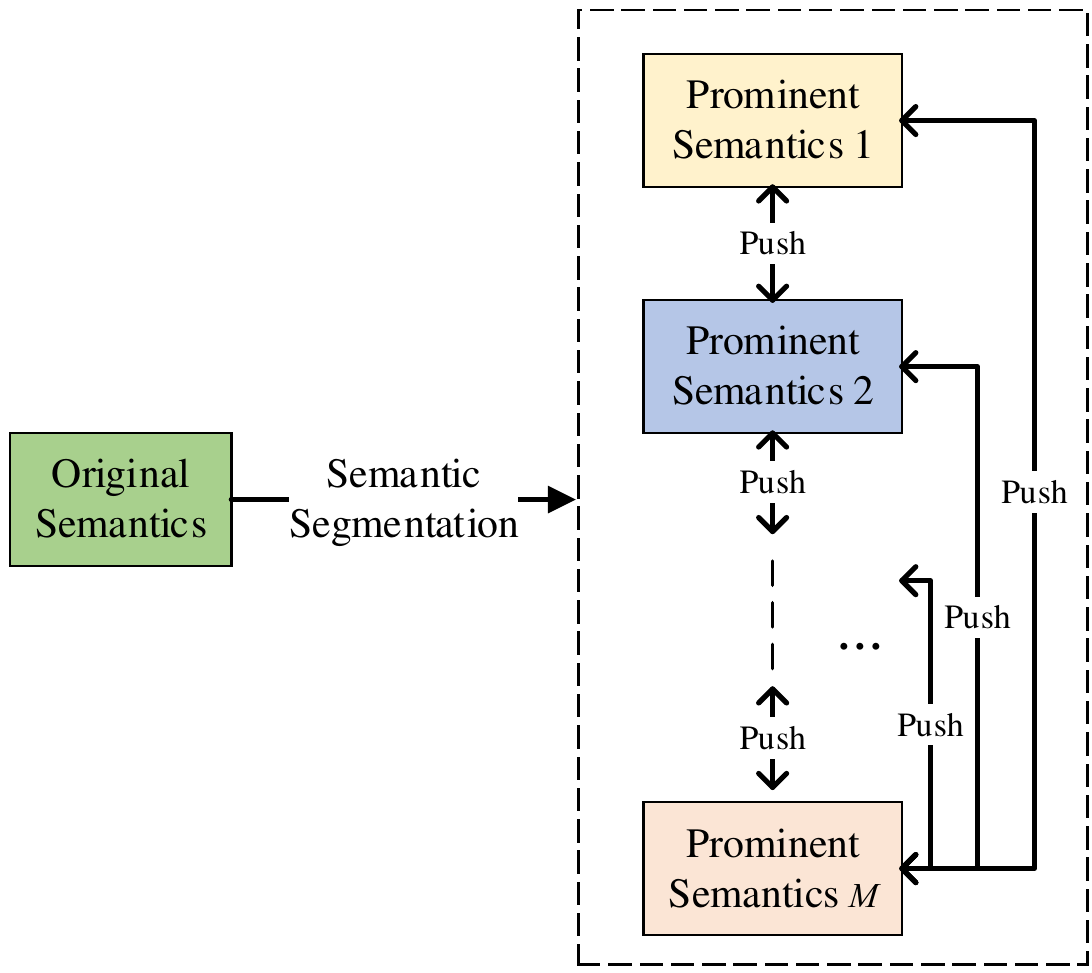}
  \caption{A Schematic of \textsc{Semantic Alienation Norm}. Note that the ``push arrow'' indicates that the semantic similarity between the Prominent Semantics at both ends is decreased.}
  \label{fig:san}
\end{figure}
We first propose $\mathcal{L}_{san}$ to make each prominent semantics as different as possible from other prominent semantics, which is computed by:
\begin{align}
\label{eq:regularization}
    \mathcal{L}_{san} &= |\mathbf{I} - \mathbf{SoftMax}(\mathbf{X} \cdot \mathbf{X}^\top)| \\
    \nonumber \mathbf{X} &= concatenate([x_{1}, x_{2}, \ldots, x_{\mathcal{M}}])
\end{align}

The SoftMax function handles the last dimension of the input matrix $\mathbf{X}$ whose shape is $\mathcal{M}\times N$.
The $\mathbf{I}$ is an identity matrix with shape $(\mathcal{M}\times \mathcal{M})$, and $x_{i}$ is the $i$-th prominent semantics vector calculated by the \Del{model's encoder }$enc$.
$\mathbf{X} \cdot \mathbf{X}^\top$ represents the correlation between a certain prominent semantic vector and other prominent semantic vectors. 
Figure~\ref{fig:san} shows a schematic of the \textsc{semantic alienation norm}.

\subsubsection{Semantic Centralization Norm}
\begin{figure}[!t]
  \centering
  \includegraphics[width=0.7\linewidth]{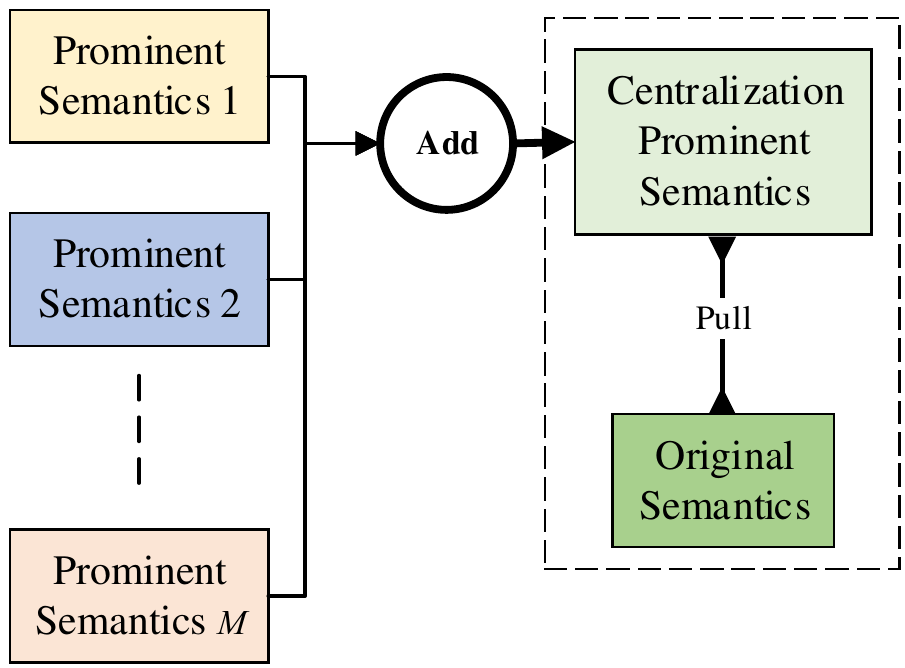}
  \caption{A Schematic of \textsc{Semantic Centralization Norm}. Note that the ``pull arrow'' indicates that the semantic similarity between the Centralization Prominent Semantics and the Original Semantics will be increased.}
  \label{fig:scn}
\end{figure}
Then we propose the $\mathcal{L}_{scn}$ to ensure the ensemble result $\sum_{i}^{\mathcal{M}}x_{i}$ of these prominent semantic vectors ($x_{1}, x_{2}, \ldots, x_{\mathcal{M}}$) is similar with the semantics of the original context, which is shown in Figure~\ref{fig:scn}. 
\begin{equation}
    \mathcal{L}_{scn} = \mathbf{1} - cosine(enc(\mathbf{C}), \sum_{i}^{\mathcal{M}}x_{i}),
\end{equation}
where $enc(\mathbf{C})$ represents the vector representation of the original semantics, $\mathbf{C}$ is the vector representation of the original context.

\subsubsection{Semantic Distillation Norm} 
\begin{figure}[!t]
  \centering
  \includegraphics[width=\linewidth]{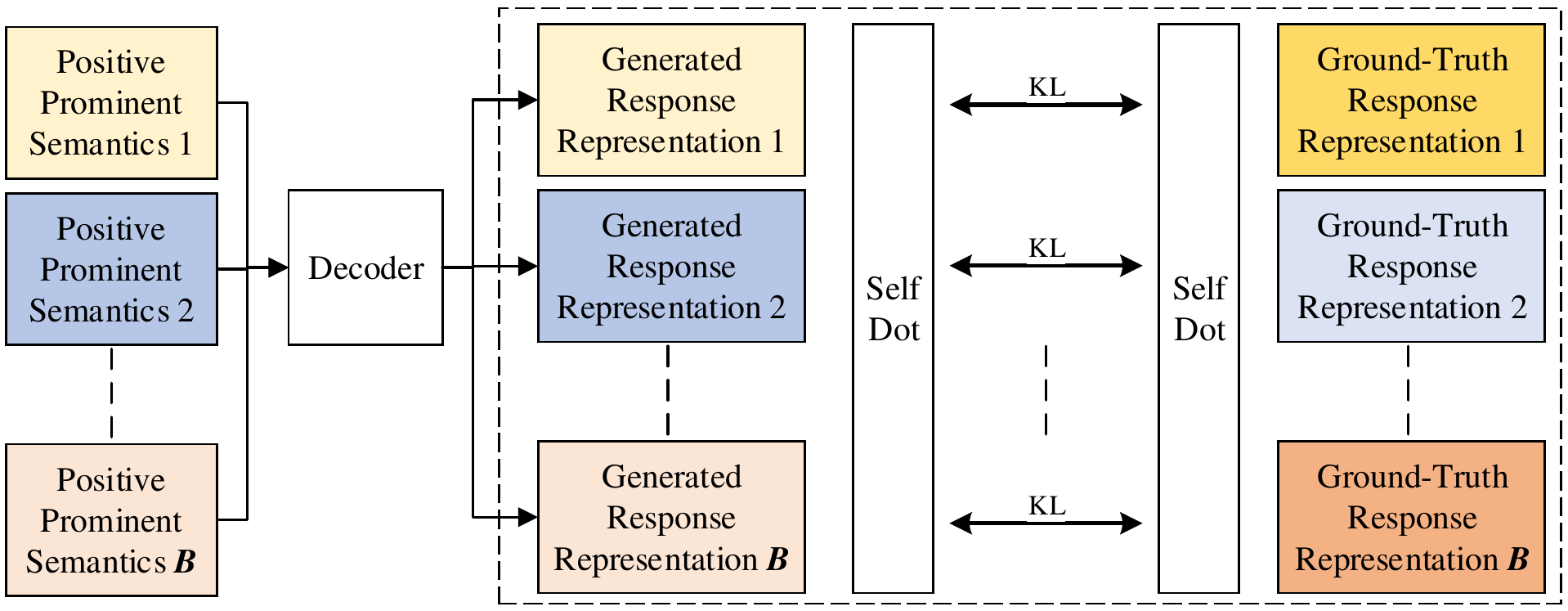}
  \caption{A Schematic of \textsc{Semantic Distillation Norm}. Note that the ``Self Dot'' operation is to make each Generated or Ground-Truth Response Representation perform an inner product with itself and other representations, and then perform SoftMax to get the correlation between each representation and all representations. KL means the KL divergence.}
  \label{fig:sdn}
\end{figure}

\begin{table*}[!t]
\small
\centering
\renewcommand\tabcolsep{3.7pt}
\begin{tabular}{@{}lccccccccc@{}}
\toprule
  Model              &ppl & Distinct-1 & Distinct-2 & Length & BLEU-1 & {BLEU-2} & BLEU-3 & Emb.Aver. & Coherence \\
  \midrule
  Seq2Seq    & 52.6$\pm$.10 & 0.006$\pm$.00 & 0.019$\pm$.00 & \ \ 6.8$\pm$.63 & 0.310$\pm$.02 & 0.243$\pm$.02 & 0.199$\pm$.02 & 0.853$\pm$.00 & 0.828$\pm$.00 \\
  CVAE   & 12.2$\pm$.13 & 0.035$\pm$.01 & 0.268$\pm$.02 & \ \ 9.7$\pm$.16 & 0.347$\pm$.00 & 0.282$\pm$.00 & 0.236$\pm$.00 & 0.842$\pm$.00 & 0.798$\pm$.00 \\
  \textbf{K}-CVAE   & \ \ 9.8$\pm$.22 & \textbf{0.045$\pm$.00} & 0.337$\pm$.01 & \ \ 9.7$\pm$.33 & 0.338$\pm$.01 & 0.275$\pm$.00 & 0.231$\pm$.00 & 0.838$\pm$.00 & 0.796$\pm$.00 \\
  SpaceFusion  & 24.3$\pm$.59  & 0.018$\pm$.00 & 0.087$\pm$.01 & \ \ 7.3$\pm$.21 & 0.335$\pm$.01 & 0.264$\pm$.01 & 0.217$\pm$.01 & 0.851$\pm$.00 & 0.825$\pm$.00 \\
  SepaCVAE     & \ \ \textbf{5.6$\pm$.07}  & 0.041$\pm$.00 & \textbf{0.367$\pm$.02} & 15.6$\pm$.75 & 0.425$\pm$.01 & 0.357$\pm$.01 & 0.306$\pm$.01 & 0.859$\pm$.00 & 0.833$\pm$.00 \\
  SegCVAE      & \ \ 6.1$\pm$.12  & 0.038$\pm$.00 & 0.341$\pm$.01 & \textbf{17.4$\pm$.27} & \textbf{0.453$\pm$.00} & \textbf{0.384$\pm$.00} & \textbf{0.330$\pm$.00} & \textbf{0.865$\pm$.00} & \textbf{0.836$\pm$.00} \\
\midrule
  Seq2Seq      & 45.9$\pm$.13 & 0.003$\pm$.00 & 0.015$\pm$.00 & 11.8$\pm$.82 & 0.236$\pm$.04 & 0.193$\pm$.03 & 0.163$\pm$.03 & 0.465$\pm$.08 & 0.281$\pm$.05 \\
  CVAE     & 12.2$\pm$.17 & 0.009$\pm$.00 & 0.131$\pm$.00 & 13.1$\pm$.24 & 0.172$\pm$.02 & 0.144$\pm$.02 & 0.123$\pm$.02 & 0.285$\pm$.04 & 0.195$\pm$.03 \\
  \textbf{K}-CVAE   & 12.1$\pm$.20 & 0.010$\pm$.00 & 0.135$\pm$.00 & 13.1$\pm$.10 & 0.202$\pm$.02 & 0.169$\pm$.02 & 0.144$\pm$.01 & 0.308$\pm$.06 & 0.198$\pm$.05 \\
  SpaceFusion    & \ \ 8.2$\pm$.02  & 0.006$\pm$.00 & 0.017$\pm$.00 & \ \ 9.7$\pm$.22 & 0.365$\pm$.01 & 0.292$\pm$.01 & 0.243$\pm$.00 & 0.808$\pm$.00 & 0.697$\pm$.00 \\
  SepaCVAE     & \ \ \textbf{2.0$\pm$.06}  & \textbf{0.025$\pm$.00} & \textbf{0.330$\pm$.03} & 13.5$\pm$.58 & 0.395$\pm$.01 & 0.326$\pm$.01 & 0.276$\pm$.01 & 0.807$\pm$.02 & 0.677$\pm$.01 \\
  SegCVAE      & \ \ 3.2$\pm$.08  & 0.021$\pm$.00 & 0.323$\pm$.01 & \textbf{14.4$\pm$.80} & \textbf{0.437$\pm$.01} & \textbf{0.364$\pm$.01} & \textbf{0.310$\pm$.01} & \textbf{0.836$\pm$.00} & \textbf{0.707$\pm$.01} \\
  \bottomrule
\end{tabular}
\caption{Results over the test data of \texttt{CornellMovie} (up) and \texttt{Opensubtitles} (down). The best score in each column is in bold. Note that our BLEU-{1,2,3} scores are normalized to [0, 1]. We run all models 5 times.}
\label{tab:dialog_general_test_results}
\end{table*}

Finally, we propose $\mathcal{L}_{sdn}$, which uses the relationship among the ground-truth responses to teach our model to learn the semantic relation of these prominent semantics. 
That is, with $\mathcal{L}_{sdn}$, the connections between prominent semantics and ground-truth responses can be further established, which can improve the consistency of response generation and the potential meaning of prominent semantics. In addition, since the representation is performed by $enc$, $\mathcal{L}_{sdn}$ can further adjust its semantic representation capability. The schematic of \textsc{semantic distillation norm} is shown in Figure~\ref{fig:sdn} and $\mathcal{L}_{sdn}$ is defined as:
\begin{align}
    \nonumber \mathcal{L}_{sdn} = KL(&\mathbf{SoftMax}(\mathbf{R}_{gt} \cdot \mathbf{R}_{gt}^\top)|| \\ &\mathbf{SoftMax}(\mathbf{R}_{gen}^{+} \cdot \mathbf{R}_{gen}^{+\top})),
\end{align}
where $\mathbf{R}_{gt}$ with the shape $(B\times N)$ represents the semantic matrix (vector representation) of batch size $B$ ground-truth responses obtained by the model's encoder $enc$.
And $\mathbf{R}_{gen}^{+}$ is the concatenated result of the vector representations of $B$ generated responses, which are obtained through the positive prominent semantics $x^+$.
Note that the SoftMax function is also used to handle the last dimension of the input matrix.

\subsection{Objective Function}
The final objective function for training our model is to maximize:
\begin{equation}
    \mathcal{L}_{all} = \mathcal{L}(r,x^{+}) - \lambda(\mathcal{L}_{san} + \mathcal{L}_{scn} + \mathcal{L}_{sdn}),
\end{equation}
where $\mathcal{L}(r,x^{+})$ is shown in Eq~\eqref{eq:L(r,Ei)}, and $\lambda$ increases linearly from 0 to 1 in the first $snorm\_step$ batches.

\section{Experiment Settings}
\label{sec:experiment}
\subsection{Data Setting}
Two well-established open domain dialogue datasets are conducted for experiment: \texttt{CornellMovie} and \texttt{Opensubtitles}.
We derived a processed version of \texttt{Opensubtitles} released by~\citet{Binsun-Sepacvae-ACL2021}, which has 5M, 100K, and 50K single-turn dialogue pairs in training, validation, and test sets, respectively.
We follow the same process for \texttt{CornellMovie} and we obtain 51,108, 6,358 and 6,249 single-turn dialogue pairs for training, validation, and test.

\subsection{Baseline Models}
\label{sect:dia-baselines}
We compare our model with state-of-the-art dialogue models: A GRU-based Seq2Seq \citep{Seq2Seq-ShangLifeng-2015,Seq2Seq-Sordoni-2015}, a general CVAE based dialogue model with BOW trick (CVAE; \citet{CVAE(SPhred)-ShenXiaoyu-2017}), a knowledge guide CVAE (\textbf{K}-CVAE; \citet{kgCVAE-ZhaoTiancheng-2017}), a SpaceFusion \citep{SpaceFusion-GaoXiang-2019} and a self-separated CVAE (SepaCVAE;  \citet{Binsun-Sepacvae-ACL2021}).
Due to the lack of knowledge annotations in datasets, we use the the cluster results of K-means (\textbf{K}) as the knowledge.

\subsection{Evaluation Metrics and Training Details}
In addition to the Distinct-n, BLEU, Emb.Aver and Coherence, we also use Perplexity (ppl) \citep{ppl} and Length \citep{FilteringData-Csaky-2019} to evaluate the performance of all models. For human evaluation, we hired three annotators to rank all models based on their generated responses. Please see Appendix~\ref{appex:exp_settings} for more details on experimental settings.

\section{Results and Analysis}

\subsection{Automatic Evaluation Results}

Table~\ref{tab:dialog_general_test_results} reports the automatic results on test data of \texttt{CornellMovie} and \texttt{Opensubtitles}. These results show that our SegCVAE achieves a better performance in terms of most metrics.
Specifically, our SegCVAE achieves the best Length, BLEU, Emb.Aver. and Coherence scores on both datasets, which demonstrates the superior performance of our model on generating coherent and related responses.
In addition, the SegCVAE has a competitive ppl and Distinct results.
Generally speaking, the Distinct metric is easily affect by the length of generated responses.
Therefore, as the SegCVAE generates longest responses, the proportion of repeated words will increase, resulting in a decrease in the distinct score.
In a nutshell, these results shows the ability of SegCVAE to handle the general dialogue generation task.

\begin{table}[!t]
\small
\centering
\renewcommand\tabcolsep{3.2pt}
\begin{tabular}{llccc}
\toprule
  DataSet & model          & Diversity & Relevance & Fluency \\
  \midrule
  & Seq2Seq                & 6.13 & 3.47 & \textbf{2.30} \\
  & CVAE               & 4.20 & 3.20 & 3.50 \\
  Cornell- & \textbf{K}-CVAE    & 2.57 & 3.33 & 3.83 \\
  Movies & SpaceFusion            & 5.13 & 3.60 & 2.73 \\
  & SepaCVAE               & 1.97 & 2.57 & 4.03 \\
  & SegCVAE                & \textbf{1.40} & \textbf{2.23} & 3.27 \\
  \cmidrule{2-5}
  & GroundTruth            & 3.60 & 1.13 & 1.03 \\
  \midrule
  & Seq2Seq                & 4.03 & 2.80 & 3.80 \\
  & CVAE               & 2.47 & 2.97 & 3.97 \\
  Open- & \textbf{K}-CVAE    & 2.73 & 3.37 & 4.00 \\
  Subtitles& SpaceFusion            & 6.57 & 2.66 & 2.30 \\
  & SepaCVAE               & 2.33 & 2.43 & 3.47 \\
  & SegCVAE                & \textbf{1.93} & \textbf{2.10} & \textbf{2.20} \\
  \cmidrule{2-5}
  & GroundTruth            & 3.93 & 1.53 & 1.07 \\
  \bottomrule
\end{tabular}
\caption{Human evaluation results on test data. The best score in each column is in bold.}
\label{tab:dialog_generation_human_eval_results}
\end{table}

\begin{table*}[!t]
\small
\centering
\renewcommand\tabcolsep{3.0pt}
\begin{tabular}{lccccccccc}

\toprule
  Model & Distinct-1 & Distinct-2 & Distinct-3 & BLEU-1 & BLEU-2 & BLEU-3 & BLEU-4 & Emb.Aver. & Coherence \\
  \midrule
  SegCVAE      & \textbf{0.021$\pm$.00} & \textbf{0.323$\pm$.01} & \textbf{0.781$\pm$.02} & \textbf{0.437$\pm$.01} & \textbf{0.364$\pm$.01} & \textbf{0.310$\pm$.01} & \textbf{0.249$\pm$.01} & \textbf{0.836$\pm$.00} & \textbf{0.707$\pm$.01} \\
  \midrule
  \ -wo. IS    & 0.010$\pm$.00 & 0.179$\pm$.02 & 0.570$\pm$.05 & 0.348$\pm$.10 & 0.291$\pm$.09 & 0.248$\pm$.07 & 0.199$\pm$.06 & 0.693$\pm$.16 & 0.519$\pm$.20 \\
  \ -wo. EG    & \textbf{0.022$\pm$.00} & \textbf{0.353$\pm$.03} & \textbf{0.816$\pm$.03} & 0.396$\pm$.02 & 0.328$\pm$.02 & 0.277$\pm$.02 & 0.222$\pm$.01 & 0.815$\pm$.01 & 0.673$\pm$.03 \\
  \ -wo. $\mathcal{L}_{san}$             & 0.018$\pm$.01 & 0.289$\pm$.07 & 0.731$\pm$.08 & 0.432$\pm$.02 & 0.358$\pm$.01 & 0.302$\pm$.01 & 0.239$\pm$.01 & \textbf{0.843$\pm$.01} & \textbf{0.727$\pm$.02}\\
  \ -wo. $\mathcal{L}_{scn}$             & 0.021$\pm$.00 & 0.313$\pm$.03 & 0.755$\pm$.05 & 0.421$\pm$.00 & 0.349$\pm$.00 & 0.296$\pm$.00 & 0.238$\pm$.00 & 0.833$\pm$.00 & 0.703$\pm$.00\\
  \ -wo. $\mathcal{L}_{sdn}$             & 0.020$\pm$.00 & 0.320$\pm$.01 & 0.774$\pm$.01 & 0.433$\pm$.00 & 0.358$\pm$.00 & 0.302$\pm$.00 & 0.243$\pm$.00 & 0.836$\pm$.01 & 0.703$\pm$.02 \\
  \bottomrule
\end{tabular}
\caption{Ablation results on test data of \texttt{Opensubtitles}. The best score in each column is in bold.} 
\label{tab:dialog_generation_ablation_results}
\end{table*}

\subsection{Human Evaluation Results}
The results of the human evaluation are shown in Table~\ref{tab:dialog_generation_human_eval_results} (refer to Appendix~\ref{sec:human_eva} for detailed setups).
To evaluate the consistency of the ranking results assessed by three annotators, we use Pearson's correlation coefficient. 
This coefficient is 0.80 on Diversity, 0.62 on Relevance, and 0.77 on Fluency, with $p < 0.0001$ and below 0.001, which indicates high correlation and agreement. 
This result shows that our model significantly outperforms baselines in terms of diversity, relevance, and fluency. 
Except for the ground-truth responses, our model achieves the best scores of relevance and diversity metrics on both datasets. 
The fluency result of SegCVAE on the \texttt{CornellMovie} is slightly worse than that of baselines, which is mainly due to the length of responses generated by SegCVAE being longer than that of baselines (see Table~\ref{tab:dialog_general_test_results}). 
When the response lengths are similar on the \texttt{Opensubtitles}, SegCVAE can also achieve the best fluency score.

\subsection{Ablation Study}
Table~\ref{tab:dialog_generation_ablation_results} reports the results of the ablation study.
It can be seen from the table that after removing IS, $\mathcal{L}_{scn}$ and $\mathcal{L}_{sdn}$, respectively, the results all decreased.
And the results decreased the most after removing IS, indicating that IS has the most important role in model performance.
In addition, we found that after removing EG, the Diversity of the model increased, but the Emb.Aver. and Coherence decreased.
This is because EG is mainly responsible for regulating the prominent semantics in the model without deviating from the original semantics.
Therefore, by removing EG, the prominent semantics obtained by IS lacks constraints and can become more diverse, but the connection with the context is weaker.
Similarly, $\mathcal{L}_{san}$ is used to make multiple prominent semantic information segmented to be different from each other, so removing $\mathcal{L}_{san}$ will reduce Diversity and increase Emb.Aver. and Coherence.

\begin{table}[!t]
\small
    \centering
    \begin{tabular}{l|p{5.5cm}}
    \toprule
        Context & I'm sorry, you're mistaken. \\
                \midrule
        EG & Confided Confided \\ 
        IS & I Mistaken \\
       SegCVAE & \textbf{So, I'll help} my mate and \textbf{you}. listen, one day to tell me to go from the fields together. \\
                \toprule
        Context & Move! What have you done? \\
                \midrule
        EG & Rendezvous Humiliate  \\ 
        IS & Move ! \\
       SegCVAE &  Hey, \textbf{please. relax}. \\
                \toprule
        Context & Not this year, dani. Mom said you have to. \\
                \midrule
        EG & Tying Tying \\ 
        IS & Said Not Said \\
       SegCVAE & I'm compounded you \textbf{talk about our great <unk> in the other times.} \\
       \bottomrule
    \end{tabular}
    \caption{Generated responses and their corresponding keyword combinations of SegCVAE. EG and IS represent the External Guidance and the Internal Separation.}
    \label{tab:premoinent_semantic_samples}
\end{table}

\subsection{Case Study}

We use the prominent semantics to guide the generation of responses, which requires the SegCVAE to learn the relations among the contexts, the prominent semantics, and the responses. 
\Del{As we analyzed above, the prominent semantics could help model the \Del{complex dialogue mappings}\Revise{CDM}, which will guide the generated responses to achieve better scores than other baseline models.}
To illustrate the connection among prominent semantics, context and generated responses, we report three samples and their related words that extract by EG and IS, which are shown in Table~\ref{tab:premoinent_semantic_samples}. Note that the words extracted by EG and IS are used for calculating prominent semantics through the encoder.

In Table~\ref{tab:premoinent_semantic_samples}, we can notice that the output of EG is difficult to relate to the response. We suppose that this would blame the poor interpretability of neural models and the lack of annotations.
Note that EG is trained by self-supervised learning without any explicit-knowledge annotations. Therefore, it learns to minimise the designed loss, which may produce some unrecognised results or intermediate features for human. We speculate that introducing annotations or knowledge that consistent with human cognition will help the model to produce more interpretable and better performance. 
We consider it as an important future work and require more efforts on this topic.

We also collect the generated responses and show them in Appendix~\ref{appex:case_study}.

\subsection{Effectiveness Analysis}
To further study the effectiveness of CDM, we conduct experiments over these mappings.

\paragraph{Data and Tasks}
We collect two particular datasets (named as \texttt{O2M} and \texttt{M2O}) from the \texttt{Opensubtitles}, and define two new tasks (one-to-many and many-to-one dialogue learning task) to analyse the ability of generative dialogue models in handling CDM. In our experiments, all models are trained on \texttt{O2M} or \texttt{M2O} to accomplish the two tasks.
The training and validation procedures are the same as for general dialogue generation task. 
In inference stage, every model should generate $N$ responses for each context in test set of \texttt{O2M} or \texttt{M2O}. Note that $N$ is set to 8 in this paper (See Appendix~\ref{appex:further_analysis} for detail).

\paragraph{Evaluation Settings} 
Different from the previous settings, we conduct a new human evaluation strategy.
First, each model received 50 contexts randomly extracted from \texttt{O2M} and \texttt{M2O}, respectively, and generated 400 responses. Then, three annotators were invited to rank all models with respect to ``Suitability'' and ``Erudition'' of their responses. Ties are allowed. Suitability indicates how many diverse and relevant responses are generated by the model. Erudition specifies whether multiple generated responses have the same semantics as the ground-truth responses.
We design Suitability to validate whether the model can learn the diversity and relevance from CDM samples, and we use Erudition to assess whether the semantic information of multiple ground-truths is involved in multiple responses generated by the model.

\begin{table}[t]
\small
\centering
\renewcommand\tabcolsep{15pt}
\begin{tabular}{lcc}
\toprule
  model          & Suitability & Erudition \\
  \midrule
  CVAE               & 2.69 & 2.33 \\
  \textbf{K}-CVAE    & 2.75 & 2.35 \\
  SepaCVAE               & 2.15 & 2.21 \\
  SegCVAE                & \textbf{2.03} & \textbf{1.89} \\
  \midrule
  CVAE               & 2.42 & 1.96 \\
  \textbf{K}-CVAE    & 2.48 & \textbf{1.89} \\
  SepaCVAE               & 2.16 & 1.92 \\
  SegCVAE                & \textbf{2.05} & 1.92 \\
  \bottomrule
\end{tabular}
\caption{Evaluation results on test data of \texttt{O2M} (up) and \texttt{M2O} (down). The best score in each column is bold.}
\label{tab:human_eval_results}
\end{table}

\paragraph{Results and Analysis} Table~\ref{tab:human_eval_results} reports the result.
We observe that SegCVAE achieves the best Suitability on both \texttt{O2M} and \texttt{M2O} datasets, which we believe stems from the model’s superior ability to model the CDM. 
We also observe that SegCVAE achieves the best Erudition on \texttt{O2M} dataset but poor Erudition on \texttt{M2O} dataset, and \textbf{K}-CVAE achieves best Erudition on \texttt{M2O} dataset but worst Erudition on \texttt{O2M} dataset. This finding is in line with the characteristics of these models:
(1) Due to the cluster information, the \textbf{K}-CVAE samples latent variables from a concentrated prior distribution, resulting in generating multiple similar responses easily. (2) The SepaCVAE uses the orthogonal vectors for sampling latent variables, which increases the diversity but decreases the number of relevant responses. (3) Our SegCVAE uses multiple prominent semantics to capture the diverse and relevant features, resulting in generating different but coherent responses.
Therefore, the similar responses generated by \textbf{K}-CVAE are easily hit the only ``one'' response in \texttt{M2O} dataset but hardly hit multiple responses in \texttt{O2M} dataset, which leads the best Erudition on \texttt{M2O} but worst Erudition on \texttt{O2M}.

On the contrary, our SegCVAE generates multiple responses corresponding to multiple prominent semantics, which easily captures the semantics of multiple responses in \texttt{O2M} dataset and achieves the best Erudition on \texttt{O2M} dataset. However, due to the trade-off between diversity and relevance, the Erudition of SegCVAE on \texttt{M2O} dataset is a little poor.
We also use the Pearson's correlation coefficient to evaluate the consistency of the ranking results. The coefficient is 0.64 on Suitability, and 0.51 on Erudition, with p<0.0001 and below 0.001, which indicates high correlation.

\section{Conclusion}
This paper proposes a novel SegCVAE to model complex dialogue mappings (CDM) in human conversations.
SegCVAE parses the CDM from a semantic perspective: Using multiple prominent semantics segmented from the context to establish relationships with the responses, multiple prominent semantics can correspond to multiple responses, and multiple contexts can also segment similar prominent semantics.
In this way, prominent semantics can constrain latent variables to learn semantic relations to tackle incoherent problem, while enriching them to mitigate the non-diverse problem.
To realize SegCVAE, we propose three novel modules: Internal Separation (IS), External Guidance (EG), and Semantic Norms (i.e. $\mathcal{L}_{san}$, $\mathcal{L}_{scn}$, and $\mathcal{L}_{sdn}$).
IS is used to get the basic information for computing prominent semantics, EG is used to constrain the prominent semantics not to deviate too far from the original semantics, and three Semantic Norms are proposed to establish relationships for contexts, prominent semantics and responses.
The experimental results show the superiority of our model in dialogue generation, one-to-many and many-to-one dialogue learning tasks.

\section*{Limitations}

The limitations of our paper are as follow: 
\begin{itemize}[align = left, wide = 1pt, itemsep=2pt, parsep=2pt, topsep = 2pt ]
    \item The SegCVAE model is proposed to model the serious complex dialogue mappings (i.e. one-to-many and many-to-one) phenomena in open-domain dialogue generation task. Therefore, the SegCVAE is suitable for generative tasks where non-one-to-one mappings exist in the dataset. If the task does not require modeling non-one-to-one mappings, our model has little advantage.
    \item The hyper-parameters (\emph{e.g.}the number of extracted words $chan$, the number of triggers $\mathcal{M}$ and so on) need to be determined through multiple experiments, which cannot be set adaptively. These initial promising results for segmenting context into multiple prominent semantics for modeling complex dialogue mappings will hopefully lead to future work in this interesting direction.
    \item We provide further analysis on One-to-Many and Many-to-One dialogue learning task, and propose a new human evaluation strategy to directly valid the performance of models on processing non-one-to-one dialogue samples. However, we do not provide results on automatic evaluation of modeling one-to-many and many-to-one mappings. This is primarily because there are no publicly recognized metrics for the evaluation of the performance on modeling one-to-many and many-to-one dialogue mappings directly. In addition, it is also difficult to propose the automatic metrics to achieve the evaluation process due to the lack of supervised information. Automatically evaluating the generative dialogue model's ability to model the complex mappings is a challenging problem and we leave that for future work.
\end{itemize}

\section*{Acknowledgements}
We would like to thank the anonymous reviewers for their constructive comments. This research is supported by Beijing Natural Science Foundation (No. 4222037 and L181010). Kan Li is the corresponding author.

\bibliography{main}

\begin{thebibliography}{44}
\expandafter\ifx\csname natexlab\endcsname\relax\def\natexlab#1{#1}\fi

\bibitem[{Akama et~al.(2020)Akama, Yokoi, Suzuki, and
  Inui}]{FilterConsistency-Akama-2020}
Reina Akama, Sho Yokoi, Jun Suzuki, and Kentaro Inui. 2020.
\newblock \href {https://doi.org/10.18653/v1/2020.emnlp-main.68} {Filtering
  noisy dialogue corpora by connectivity and content relatedness}.
\newblock In \emph{{EMNLP}}, pages 941--958.

\bibitem[{Chen et~al.(2018)Chen, Ren, Tang, Zhao, and
  Yin}]{HVaeMN-ChenHongshen-2018}
Hongshen Chen, Zhaochun Ren, Jiliang Tang, Yihong~Eric Zhao, and Dawei Yin.
  2018.
\newblock \href {https://doi.org/10.1145/3178876.3186077} {Hierarchical
  variational memory network for dialogue generation}.
\newblock In \emph{{WWW}}, pages 1653--1662. {ACM}.

\bibitem[{Csaky et~al.(2019)Csaky, Purgai, and
  Recski}]{FilteringData-Csaky-2019}
Richard Csaky, Patrik Purgai, and G{\'{a}}bor Recski. 2019.
\newblock \href {https://doi.org/10.18653/v1/p19-1567} {Improving neural
  conversational models with entropy-based data filtering}.
\newblock In \emph{{ACL} {(1)}}, pages 5650--5669.

\bibitem[{Danescu-Niculescu-Mizil and Lee(2011)}]{cornellmovie}
Cristian Danescu-Niculescu-Mizil and Lillian Lee. 2011.
\newblock \href {https://www.aclweb.org/anthology/W11-0609/} {Chameleons in
  imagined conversations: A new approach to understanding coordination of
  linguistic style in dialogs.}
\newblock In \emph{{ACL}}.

\bibitem[{Feng et~al.(2020{\natexlab{a}})Feng, Chen, Li, and
  Yin}]{PosteriorGan-FengShaoxiong2020}
Shaoxiong Feng, Hongshen Chen, Kan Li, and Dawei Yin. 2020{\natexlab{a}}.
\newblock \href {https://aaai.org/ojs/index.php/AAAI/article/view/6273}
  {Posterior-gan: Towards informative and coherent response generation with
  posterior generative adversarial network}.
\newblock In \emph{{AAAI}}, pages 7708--7715.

\bibitem[{Feng et~al.(2020{\natexlab{b}})Feng, Ren, Chen, Sun, Li, and
  Sun}]{Regularizing-Feng-2020}
Shaoxiong Feng, Xuancheng Ren, Hongshen Chen, Bin Sun, Kan Li, and Xu~Sun.
  2020{\natexlab{b}}.
\newblock \href {https://doi.org/10.18653/v1/2020.emnlp-main.534} {Regularizing
  dialogue generation by imitating implicit scenarios}.
\newblock In \emph{{EMNLP}}, pages 6592--6604.

\bibitem[{Gao et~al.(2019{\natexlab{a}})Gao, Bi, Liu, Li, Zhou, and
  Shi}]{discrete-cvae-2019-emnlp}
Jun Gao, Wei Bi, Xiaojiang Liu, Junhui Li, Guodong Zhou, and Shuming Shi.
  2019{\natexlab{a}}.
\newblock \href {https://doi.org/10.18653/v1/D19-1198} {A discrete {CVAE} for
  response generation on short-text conversation}.
\newblock In \emph{{EMNLP-IJCNLP}}, pages 1898--1908. Association for
  Computational Linguistics.

\bibitem[{Gao et~al.(2019{\natexlab{b}})Gao, Lee, Zhang, Brockett, Galley, Gao,
  and Dolan}]{SpaceFusion-GaoXiang-2019}
Xiang Gao, Sungjin Lee, Yizhe Zhang, Chris Brockett, Michel Galley, Jianfeng
  Gao, and Bill Dolan. 2019{\natexlab{b}}.
\newblock \href {https://doi.org/10.18653/v1/n19-1125} {Jointly optimizing
  diversity and relevance in neural response generation}.
\newblock In \emph{{NAACL-HLT} {(1)}}, pages 1229--1238.

\bibitem[{Ghazvininejad et~al.(2018)Ghazvininejad, Brockett, Chang, Dolan, Gao,
  Yih, and Galley}]{FactKnowledge-Ghazvininejad-2018}
Marjan Ghazvininejad, Chris Brockett, Ming{-}Wei Chang, Bill Dolan, Jianfeng
  Gao, Wen{-}tau Yih, and Michel Galley. 2018.
\newblock \href
  {https://www.aaai.org/ocs/index.php/AAAI/AAAI18/paper/view/16710} {A
  knowledge-grounded neural conversation model}.
\newblock In \emph{{AAAI}}, pages 5110--5117.

\bibitem[{He and Glass(2020)}]{NegativeTrain-He-2020}
Tianxing He and James~R. Glass. 2020.
\newblock \href {https://doi.org/10.18653/v1/2020.acl-main.185} {Negative
  training for neural dialogue response generation}.
\newblock In \emph{ACL}, pages 2044--2058.

\bibitem[{Huber et~al.(2018)Huber, McDuff, Brockett, Galley, and
  Dolan}]{EmotionalTextImage-Huber-2018}
Bernd Huber, Daniel~J. McDuff, Chris Brockett, Michel Galley, and Bill Dolan.
  2018.
\newblock \href {https://doi.org/10.1145/3173574.3173851} {Emotional dialogue
  generation using image-grounded language models}.
\newblock In \emph{{CHI}}, page 277.

\bibitem[{Jang et~al.(2017)Jang, Gu, and Poole}]{gumbel-softmax-17}
Eric Jang, Shixiang Gu, and Ben Poole. 2017.
\newblock \href {https://openreview.net/forum?id=rkE3y85ee} {Categorical
  reparameterization with gumbel-softmax}.
\newblock In \emph{{ICLR} (Poster)}. OpenReview.net.

\bibitem[{Kingma and
  Welling(2014)}]{StochasticGradientVariationalBayes-kingma-2014}
Diederik~P. Kingma and Max Welling. 2014.
\newblock \href {http://arxiv.org/abs/1312.6114} {Auto-encoding variational
  bayes}.
\newblock In \emph{{ICLR}}.

\bibitem[{Li et~al.(2016{\natexlab{a}})Li, Galley, Brockett, Gao, and
  Dolan}]{distinct-16}
Jiwei Li, Michel Galley, Chris Brockett, Jianfeng Gao, and Bill Dolan.
  2016{\natexlab{a}}.
\newblock \href {https://doi.org/10.18653/v1/n16-1014} {A diversity-promoting
  objective function for neural conversation models}.
\newblock In \emph{{HLT-NAACL}}, pages 110--119.

\bibitem[{Li et~al.(2016{\natexlab{b}})Li, Galley, Brockett, Spithourakis, Gao,
  and Dolan}]{Persona-LiJiwei-2016}
Jiwei Li, Michel Galley, Chris Brockett, Georgios~P. Spithourakis, Jianfeng
  Gao, and William~B. Dolan. 2016{\natexlab{b}}.
\newblock \href {https://doi.org/10.18653/v1/p16-1094} {A persona-based neural
  conversation model}.
\newblock In \emph{{ACL} {(1)}}.

\bibitem[{Li et~al.(2016{\natexlab{c}})Li, Monroe, Ritter, Jurafsky, Galley,
  and Gao}]{RLdialoguesys-LiJiwei-2016}
Jiwei Li, Will Monroe, Alan Ritter, Dan Jurafsky, Michel Galley, and Jianfeng
  Gao. 2016{\natexlab{c}}.
\newblock \href {https://doi.org/10.18653/v1/d16-1127} {Deep reinforcement
  learning for dialogue generation}.
\newblock In \emph{{EMNLP}}, pages 1192--1202.

\bibitem[{Li et~al.(2022{\natexlab{a}})Li, Feng, Sun, and
  Li}]{NegDistill-yiweili-naacl-2022}
Yiwei Li, Shaoxiong Feng, Bin Sun, and Kan Li. 2022{\natexlab{a}}.
\newblock \href {https://doi.org/10.18653/v1/2022.naacl-main.31} {Diversifying
  neural dialogue generation via negative distillation}.
\newblock In \emph{Proceedings of the 2022 Conference of the North American
  Chapter of the Association for Computational Linguistics: Human Language
  Technologies, {NAACL} 2022, Seattle, WA, United States, July 10-15, 2022},
  pages 407--418. Association for Computational Linguistics.

\bibitem[{Li et~al.(2022{\natexlab{b}})Li, Sun, Feng, and
  Li}]{StopFiltering-yiweili-2022}
Yiwei Li, Bin Sun, Shaoxiong Feng, and Kan Li. 2022{\natexlab{b}}.
\newblock \href {https://doi.org/10.48550/arXiv.2205.11206} {Stop filtering:
  Multi-view attribute-enhanced dialogue learning}.
\newblock \emph{CoRR}, abs/2205.11206.

\bibitem[{Lison and Tiedemann(2016)}]{opensubtitles2016}
Pierre Lison and J{\"{o}}rg Tiedemann. 2016.
\newblock \href
  {http://www.lrec-conf.org/proceedings/lrec2016/summaries/947.html}
  {Opensubtitles2016: Extracting large parallel corpora from movie and {TV}
  subtitles}.
\newblock In \emph{{LREC}}.

\bibitem[{Liu et~al.(2016)Liu, Lowe, Serban, Noseworthy, Charlin, and
  Pineau}]{embedding-16}
Chia{-}Wei Liu, Ryan Lowe, Iulian Serban, Michael Noseworthy, Laurent Charlin,
  and Joelle Pineau. 2016.
\newblock \href {https://doi.org/10.18653/v1/d16-1230} {How {NOT} to evaluate
  your dialogue system: An empirical study of unsupervised evaluation metrics
  for dialogue response generation}.
\newblock In \emph{{EMNLP}}, pages 2122--2132.

\bibitem[{Liu et~al.(2020)Liu, Chen, Chen, Lou, Chen, Zhou, and
  Zhang}]{RL-P2BOT-Liu2020}
Qian Liu, Yihong Chen, Bei Chen, Jian{-}Guang Lou, Zixuan Chen, Bin Zhou, and
  Dongmei Zhang. 2020.
\newblock \href {https://doi.org/10.18653/v1/2020.acl-main.131} {You impress
  me: Dialogue generation via mutual persona perception}.
\newblock In \emph{{ACL}}, pages 1417--1427.

\bibitem[{Luong et~al.(2015)Luong, Pham, and
  Manning}]{Attention-ThangLuong-2015}
Thang Luong, Hieu Pham, and Christopher~D. Manning. 2015.
\newblock \href {https://doi.org/10.18653/v1/d15-1166} {Effective approaches to
  attention-based neural machine translation}.
\newblock In \emph{{EMNLP}}, pages 1412--1421.

\bibitem[{Mi et~al.(2022)Mi, Li, Zeng, Zhou, Wang, Xu, Shang, Jiang, Zhao, and
  Liu}]{PanGuBot-feimi-2022}
Fei Mi, Yitong Li, Yulong Zeng, Jingyan Zhou, Yasheng Wang, Chuanfei Xu, Lifeng
  Shang, Xin Jiang, Shiqi Zhao, and Qun Liu. 2022.
\newblock \href {https://doi.org/10.48550/arXiv.2203.17090} {{PANGUBOT:}
  efficient generative dialogue pre-training from pre-trained language model}.
\newblock \emph{CoRR}, abs/2203.17090.

\bibitem[{Neubig(2017)}]{ppl}
Graham Neubig. 2017.
\newblock \href {http://arxiv.org/abs/1703.01619} {Neural machine translation
  and sequence-to-sequence models: {A} tutorial}.
\newblock \emph{CoRR}, abs/1703.01619.

\bibitem[{Papineni et~al.(2002)Papineni, Roukos, Ward, and Zhu}]{bleu}
Kishore Papineni, Salim Roukos, Todd Ward, and Wei{-}Jing Zhu. 2002.
\newblock \href {https://doi.org/10.3115/1073083.1073135} {Bleu: a method for
  automatic evaluation of machine translation}.
\newblock In \emph{{ACL}}, pages 311--318.

\bibitem[{Pennington et~al.(2014)Pennington, Socher, and
  Manning}]{GloVe-Pennington-2014}
Jeffrey Pennington, Richard Socher, and Christopher~D. Manning. 2014.
\newblock \href {https://doi.org/10.3115/v1/d14-1162} {Glove: Global vectors
  for word representation}.
\newblock In \emph{{EMNLP}}, pages 1532--1543.

\bibitem[{Serban et~al.(2016)Serban, Sordoni, Bengio, Courville, and
  Pineau}]{HRED-Dialogue-Serban-2016}
Iulian~Vlad Serban, Alessandro Sordoni, Yoshua Bengio, Aaron~C. Courville, and
  Joelle Pineau. 2016.
\newblock \href
  {http://www.aaai.org/ocs/index.php/AAAI/AAAI16/paper/view/11957} {Building
  end-to-end dialogue systems using generative hierarchical neural network
  models}.
\newblock In \emph{{AAAI}}, pages 3776--3784.

\bibitem[{Shang et~al.(2015)Shang, Lu, and Li}]{Seq2Seq-ShangLifeng-2015}
Lifeng Shang, Zhengdong Lu, and Hang Li. 2015.
\newblock \href {https://doi.org/10.3115/v1/p15-1152} {Neural responding
  machine for short-text conversation}.
\newblock In \emph{{ACL} {(1)}}, pages 1577--1586.

\bibitem[{Shen et~al.(2017)Shen, Su, Li, Li, Niu, Zhao, Aizawa, and
  Long}]{CVAE(SPhred)-ShenXiaoyu-2017}
Xiaoyu Shen, Hui Su, Yanran Li, Wenjie Li, Shuzi Niu, Yang Zhao, Akiko Aizawa,
  and Guoping Long. 2017.
\newblock \href {https://doi.org/10.18653/v1/P17-2080} {A conditional
  variational framework for dialog generation}.
\newblock In \emph{{ACL} {(2)}}, pages 504--509.

\bibitem[{Sohn et~al.(2015)Sohn, Lee, and Yan}]{ELBO1-Sohn-2015}
Kihyuk Sohn, Honglak Lee, and Xinchen Yan. 2015.
\newblock \href
  {https://proceedings.neurips.cc/paper/2015/hash/8d55a249e6baa5c06772297520da2051-Abstract.html}
  {Learning structured output representation using deep conditional generative
  models}.
\newblock In \emph{{NIPS}}, pages 3483--3491.

\bibitem[{Sordoni et~al.(2015)Sordoni, Galley, Auli, Brockett, Ji, Mitchell,
  Nie, Gao, and Dolan}]{Seq2Seq-Sordoni-2015}
Alessandro Sordoni, Michel Galley, Michael Auli, Chris Brockett, Yangfeng Ji,
  Margaret Mitchell, Jian{-}Yun Nie, Jianfeng Gao, and Bill Dolan. 2015.
\newblock \href {https://doi.org/10.3115/v1/n15-1020} {A neural network
  approach to context-sensitive generation of conversational responses}.
\newblock In \emph{{HLT-NAACL}}, pages 196--205.

\bibitem[{Sun et~al.(2021)Sun, Feng, Li, Liu, and Li}]{Binsun-Sepacvae-ACL2021}
Bin Sun, Shaoxiong Feng, Yiwei Li, Jiamou Liu, and Kan Li. 2021.
\newblock \href {https://doi.org/10.18653/v1/2021.acl-long.437} {Generating
  relevant and coherent dialogue responses using self-separated conditional
  variational autoencoders}.
\newblock In \emph{{ACL/IJCNLP}}, pages 5624--5637. ACL.

\bibitem[{Sun et~al.(2022)Sun, Feng, Li, Liu, and Li}]{THINK-binsun-2022}
Bin Sun, Shaoxiong Feng, Yiwei Li, Jiamou Liu, and Kan Li. 2022.
\newblock \href {https://doi.org/10.1016/j.knosys.2022.108376} {{THINK:} {A}
  novel conversation model for generating grammatically correct and coherent
  responses}.
\newblock \emph{Knowl. Based Syst.}, 242:108376.

\bibitem[{Sutskever et~al.(2014)Sutskever, Vinyals, and
  Le}]{Seq2Seq-Sutskever-2014}
Ilya Sutskever, Oriol Vinyals, and Quoc~V. Le. 2014.
\newblock \href
  {https://proceedings.neurips.cc/paper/2014/hash/a14ac55a4f27472c5d894ec1c3c743d2-Abstract.html}
  {Sequence to sequence learning with neural networks}.
\newblock In \emph{{NIPS}}, pages 3104--3112.

\bibitem[{Tao et~al.(2018)Tao, Gao, Shang, Wu, Zhao, and
  Yan}]{CMHAM-TaoChongyang-2018}
Chongyang Tao, Shen Gao, Mingyue Shang, Wei Wu, Dongyan Zhao, and Rui Yan.
  2018.
\newblock \href {https://doi.org/10.24963/ijcai.2018/614} {Get the point of my
  utterance! learning towards effective responses with multi-head attention
  mechanism}.
\newblock In \emph{{IJCAI}}, pages 4418--4424.

\bibitem[{Wang et~al.(2019)Wang, Feng, Wang, and
  Zhang}]{DBLP:conf/emnlp/WangFWZ19}
Weichao Wang, Shi Feng, Daling Wang, and Yifei Zhang. 2019.
\newblock \href {https://doi.org/10.18653/v1/D19-1511} {Answer-guided and
  semantic coherent question generation in open-domain conversation}.
\newblock In \emph{EMNLP-IJCNLP}, pages 5065--5075. Association for
  Computational Linguistics.

\bibitem[{Xu et~al.(2018{\natexlab{a}})Xu, Ren, Lin, and
  Sun}]{DPGAN-XuJingjing-2018}
Jingjing Xu, Xuancheng Ren, Junyang Lin, and Xu~Sun. 2018{\natexlab{a}}.
\newblock \href {https://doi.org/10.18653/v1/d18-1428} {Diversity-promoting
  {GAN:} {A} cross-entropy based generative adversarial network for diversified
  text generation}.
\newblock In \emph{{EMNLP}}, pages 3940--3949.

\bibitem[{Xu et~al.(2018{\natexlab{b}})Xu, Dusek, Konstas, and
  Rieser}]{FilterCoherence-Xu-2018}
Xinnuo Xu, Ondrej Dusek, Ioannis Konstas, and Verena Rieser.
  2018{\natexlab{b}}.
\newblock \href {https://doi.org/10.18653/v1/d18-1432} {Better conversations by
  modeling, filtering, and optimizing for coherence and diversity}.
\newblock In \emph{{EMNLP}}, pages 3981--3991.

\bibitem[{Xu et~al.(2018{\natexlab{c}})Xu, Dusek, Konstas, and
  Rieser}]{coherence}
Xinnuo Xu, Ondrej Dusek, Ioannis Konstas, and Verena Rieser.
  2018{\natexlab{c}}.
\newblock \href {https://doi.org/10.18653/v1/d18-1432} {Better conversations by
  modeling, filtering, and optimizing for coherence and diversity}.
\newblock In \emph{{EMNLP}}, pages 3981--3991.

\bibitem[{Xu et~al.(2017)Xu, Liu, Wang, Sun, Wang, Wang, and
  Qi}]{GAN-GANAEL-Xu2017}
Zhen Xu, Bingquan Liu, Baoxun Wang, Chengjie Sun, Xiaolong Wang, Zhuoran Wang,
  and Chao Qi. 2017.
\newblock \href {https://doi.org/10.18653/v1/d17-1065} {Neural response
  generation via {GAN} with an approximate embedding layer}.
\newblock In \emph{{EMNLP}}, pages 617--626.

\bibitem[{Yan et~al.(2016)Yan, Yang, Sohn, and Lee}]{ELBO2-Yan-2016}
Xinchen Yan, Jimei Yang, Kihyuk Sohn, and Honglak Lee. 2016.
\newblock \href {https://doi.org/10.1007/978-3-319-46493-0\_47}
  {Attribute2image: Conditional image generation from visual attributes}.
\newblock In \emph{{ECCV} {(4)}}, volume 9908 of \emph{Lecture Notes in
  Computer Science}, pages 776--791.

\bibitem[{Zhang et~al.(2018{\natexlab{a}})Zhang, Lan, Guo, Xu, and
  Cheng}]{RL-Seq2seqCo-Zhang2018}
Hainan Zhang, Yanyan Lan, Jiafeng Guo, Jun Xu, and Xueqi Cheng.
  2018{\natexlab{a}}.
\newblock \href {https://doi.org/10.24963/ijcai.2018/635} {Reinforcing
  coherence for sequence to sequence model in dialogue generation}.
\newblock In \emph{{IJCAI}}, pages 4567--4573.

\bibitem[{Zhang et~al.(2018{\natexlab{b}})Zhang, Galley, Gao, Gan, Li,
  Brockett, and Dolan}]{ConverseGAN(AIM)-ZhangYizhe-2018}
Yizhe Zhang, Michel Galley, Jianfeng Gao, Zhe Gan, Xiujun Li, Chris Brockett,
  and Bill Dolan. 2018{\natexlab{b}}.
\newblock \href
  {https://proceedings.neurips.cc/paper/2018/hash/23ce1851341ec1fa9e0c259de10bf87c-Abstract.html}
  {Generating informative and diverse conversational responses via adversarial
  information maximization}.
\newblock In \emph{NeurIPS}, pages 1815--1825.

\bibitem[{Zhao et~al.(2017)Zhao, Zhao, and
  Esk{\'{e}}nazi}]{kgCVAE-ZhaoTiancheng-2017}
Tiancheng Zhao, Ran Zhao, and Maxine Esk{\'{e}}nazi. 2017.
\newblock \href {https://doi.org/10.18653/v1/P17-1061} {Learning
  discourse-level diversity for neural dialog models using conditional
  variational autoencoders}.
\newblock In \emph{{ACL} {(1)}}, pages 654--664.

\end{thebibliography}
\bibliographystyle{acl_natbib}

\newpage
\appendix

\section{Experimental Settings}
\label{appex:exp_settings}
\subsection{Automatic Evaluation Metrics}
\label{sect:dia-met}
Our primary metrics of interest are Distinct-n, BLEU, Embedding Average (Emb.Aver.), and Coherence. The Distinct-n is responsible for evaluating the diversity of generated responses, which is calculated through the ratio of distinct $n$-grams and all generated $n$-grams.
The BLEU is used to evaluate the degree of the word-overlap between the generated response and the ground truth response.
The Emb.Aver. is introduced to evaluate the semantic relationship of generated responses and ground-truth responses. 
The Coherence is applied to assess the coherence and relevance between contexts and generated responses.
In addition, we also employ the Perplexity (ppl) and Length to validate all models. The ppl is an indicator commonly used in dialogue generation tasks, which is usually used to assess the degree of convergence of the model. The Response length is the average number of words of all generated responses.

\subsection{Human Evaluation}
\label{sec:human_eva}
We conduct a human evaluation to further validate our model and baseline models for their ability to generate proper responses.
First, we randomly extracted 200 samples from the test sets of the two dialogue datasets, respectively. Each sample contains one context and the response generated by different models.
Then, we hired three annotators to rank all responses of all experimented models concerning three aspects of the generated responses: Diversity, Relevance and Fluency. Ties are allowed. Diversity indicates how much the generated response provides specific and diverse information. The more diverse the information, the higher the ranking of the model on Diversity. Relevance specifies how likely the generated response is relevant to the context, which requires annotators to consider whether the response is suitable for the context at a semantic level. Fluency represents how likely the generated response is produced by human. The fewer syntactic errors, the higher the model will rank in terms of Fluency.

\begin{table}[!t]
    \small
    \centering
    \begin{tabular}{lp{5.2cm}}
    \toprule
        Context & That's what your crew said. Are \textbf{they lying}? or are you? \\
    \midrule
        Seq2Seq & I'm not going to be a long time. \\ 
       CVAE & Great. You've brought us before that's ten o'clock. \\ 
       \textbf{K}-CVAE & You played the maximum? \\ 
       SpaceFusion & I'm not a couple of firing.  \\ 
        SepaCVAE & These're what ... to now you? \\
        SegCVAE & Well, We know it's twenty, new york, \textbf{they're throwing the company.} \\
    \midrule
        Context & We \textbf{don't know.} \\
    \midrule
        Seq2Seq & I don't know. \\ 
       CVAE & I told you, Mrs. Webster. You don't wanna put the police and he's dead. \\ 
       \textbf{K}-CVAE & Now it's something i want to know that. \\ 
       SpaceFusion & What do you think they were going to do?  \\ 
       SepaCVAE & Do you see of what stupid well here? \\
        SegCVAE & What the matter? \textbf{You think we understand anything to that, anything?} \\ 
    \bottomrule
    \end{tabular}
    \caption{Generated responses from the baseline and
SegCVAE on test set of \texttt{CornellMovie}.}
    \label{tab:dialog_generation_cases}
\end{table}

\begin{table*}[ht]
\small
\centering
\renewcommand\tabcolsep{5.1pt}
\begin{tabular}{llllllllll}
\toprule
    dataset    &    type & \# tokens  & \# pairs & \# contexts($c$) & \# responses($r$) & avg \# $r$ & avg \# $c$ & max \# $r$ & max \# $c$\\ 
    \midrule
        & training   & 40,875  & 778,658   & 284,516   & 778,658   & 2.74 & -    & 1,546 & -     \\
    \texttt{O2M} & validation & -       & 222,126   & 81,057    & 222,126   & 2.74 & -    & 689   & -     \\
        & test       & -       & 110,446   & 40,710    & 110,446   & 2.71 & -    & 497   & -     \\
        \hline
        & training   & 40,331  & 768,183   & 768,183   & 279,978   & -    & 2.74 & -     & 1,588 \\
    \texttt{M2O} & validation & -       & 217,474   & 217,474   & 79,552    & -    & 2.73 & -     & 957   \\
        & test       & -       & 109,815   & 109,815   & 39,795    & -    & 2.76 & -     & 321   \\
\bottomrule
\end{tabular}
\caption{Statistics for \texttt{One-to-Many} (\texttt{O2M}) and \texttt{Many-to-One} (\texttt{M2O}) datasets. The \# tokens is the vocabulary size, and the \# pairs/contexts/responses is the number of the dialogue pairs/contexts/responses in datasets. The avg/max \# r is the average/maximum number of responses for each context, and the avg/max \# c is the average/maximum number of contexts for each response. ``-'' means the cell is not necessary for this type/dataset.}
\label{tab:data_statistics}
\end{table*}

\subsection{Training Details}
\label{sec:training_details}
For fair comparison, we used the 300-dimensional GloVe embeddings as the word-embedding matrix for all models. The hidden size of all models is set to 300. The maximum length of context and response is set to 25. The $m$, $chan$ and $\mathcal{M}$ are set to 3, 3 and 8, respectively. We set the batch sizes to 64 and 32 for \texttt{CornellMovie} and \texttt{Opensubtitles}, respectively. Adam is utilized for optimization. The initial learning rate is set to 0.001.
The $snorm_step$ is set to 20000 for \texttt{CornellMovie}, but for \texttt{Opensubtitles}, the $\lambda$ is constant at 1.0. We also introduce KL annealing trick to leverage the KL divergence during the training. The KL weight increases linearly from 0 to 1 in the first 10000 batches.
We train all models in 50 epochs on a RTX 2080Ti GPU card with Tensorflow, and save the generated responses when the ppl reaching minimum. The random seed is set as 123456. Greedy search is used to generate responses for evaluation.

\section{Case Study}
\label{appex:case_study}

We collected the generated responses from the test set of \texttt{CornellMovie} and showed them in Table~\ref{tab:dialog_generation_cases}. 
In the first example, we found that SegCVAE gave a response of ``they're throwing the company.'' considering ``they lying'' in the context. Compared with the responses generated by other models, the response of SegCVAE is more specific and more relevant to the context.
As for the second sample, only the Seq2seq only generates a general and short reply "I don't know."; the others all generate diverse responses. However, considering the coherence between the generated responses and the context, our model is more advantageous. This result shows the superiority of SegCVAE in solving the dialogue context and generating diverse response for dialogue generation task. As we have analyzed, using the prominent semantics to replace the original context for response generation can better establish the semantic relationship between context and response, thus ensuring the diversity and relevance of the generated responses.

\section{Further Analysis on One-to-Many and Many-to-One Dialogue Learning}
\label{appex:further_analysis}

\subsection{Data Settings}
We extract two particular datasets from the raw \texttt{Opensubtitles}: One-to-Many and Many-to-One, for the One-to-Many and Many-to-One dialogue learning, respectively. To build these two datasets, we first extract single-turn dialogues from the \texttt{Opensubtitles}: $T-1$ single-turn dialogues $[(u_1,u_2),(u_2,u_3),...,(u_{T-1},u_{T})]$ can be extracted from one multi-turn dialogue $(u_1,u_2,...,u_T)$, where $u$ represents an utterance in each dialogue. Then, we selected and collected a large collection of one-to-many dialogue pairs as the \texttt{One-to-Many} (\texttt{O2M}) dataset, and another large collection of many-to-one dialogue pairs as the \texttt{Many-to-One} (\texttt{M2O}) dataset. Finally, we use the token-list of GloVe \citep{GloVe-Pennington-2014} to filter the \texttt{O2M} and \texttt{M2O} datasets. For each dialogue pair (context $c_i$, response $r_i$), we first obtain its tokens after word segmentation, and then judge whether its tokens are all contained in GloVe's token-list. If the GloVe do not contain any tokens of ($c_i$, $r_i$), we drop all dialogue pairs containing the $c_i$ or $r_i$ from the dataset.
Table~\ref{tab:data_statistics} lists key statistics of the dataset after processing.

\subsection{Non-one-to-one Dialogue Learning Tasks}
\paragraph{One-to-Many Dialogue Learning Task} 
Let $c$ denote a context, and $rs$={$r_1, r_2, \ldots, r_n$} denote the responses to $c$. Follow the general dialogue generation task, we put the $c$ and $rs$ into $n$ dialogue pairs $(c, r_1), (c, r_2), \ldots, (c, r_n)$. Let $\mathcal{D}_{1n}$ represent the dataset that only contains such one-to-many dialogue pairs. This task requires a dialogue generation model to learn the one-to-many knowledge, and to generate multiple coherent and informative responses for every context sentence.

\begin{table}[!t]
\small
    \centering
    \begin{tabular}{l|p{5.4cm}}
    \toprule
        Context & I'd rather die than live with you! freaking unk! \\
                \midrule
      Responses & \textbf{Relax!} where does it hurt? \\ 
                & \textbf{Stop!} ma'am, ma'am! \\
                \midrule
       CVAE & I'm gonna get you to know! \\ 
                & That's a bad idea, mister. \\ 
                & I have a hell! \\ 
                & It's a joke that you said he's a special agent! why do you want me to believe? \\ 
                & You have something to do with this? aah. \\ 
                & Hey, you're ready? yeah. \\ 
                & The world's in the mood! \\ 
                & Here, put your hands in the bowl.\\
                \midrule
        SegCVAE & Yep tonight really... to me. sean? \\ 
                & \textbf{Calm down.} hurry any, hurry unk. \\ 
                & Nothing, they are hot / hey, \\ 
                & \textbf{No-no,} your unk. i... \\ 
                & God? uh... did not fit... \\ 
                & Be it then let's abandon it. 9 pigs. 1 50,000. open. \\ 
                & Really is going with nothing? all unk came in the past hours. \\ 
                & Most way. hell and i are unk \\
                \bottomrule
    \end{tabular}
     \caption{Generated responses from the baseline and
SegCVAE on \texttt{O2M} dataset.}
    \label{tab:o2m_samples}
\end{table}

\paragraph{Many-to-One Dialogue Learning Task} 
Relatively speaking, let $cs$={$c_1, c_2, \ldots, c_n$} denote the contexts, and $r$ denote a response to the $cs$. Correspondingly, we use $\mathcal{D}_{n1}$ to represent a dataset that only contains many-to-one dialogue pairs $(c_1, r), (c_2, r), \ldots, (c_n, r)$. This task requires the dialogue generation model to learn the many-to-one knowledge, and to distinguish which of the contexts can give the same response, and then increase the diversity while keeping the coherence of the generated response.

In our experiments, all models are trained on $\mathcal{D}_{1n}$ or $\mathcal{D}_{n1}$ to accomplish the One-to-Many Dialogue Learning Task or Many-to-One Dialogue Learning Task. The training and validation procedures are the same as for general dialogue generation task. In inference stage, every model should generate $N$ responses for each context in test set of $\mathcal{D}_{1n}$ or $\mathcal{D}_{n1}$. Note that $N$ is set to 8 in this paper. 

\subsection{Case Study}
We collected the generated responses of contexts in test set of \texttt{O2M} dataset and showed a sample in Table~\ref{tab:o2m_samples}.
We can observe that the SegCVAE generates ``Calm down.'' and ``No-no,'', which are corresponding to the ``Relax!'' and ``Stop'' in true responses. 
This result illustrates that the SegCVAE can effectively build the relations between the multiple prominent semantics and the multiple responses.

\end{document}